\documentclass[runningheads]{llncs}

\usepackage{graphicx}
\usepackage{multirow}
\usepackage{svg}
\usepackage[misc]{ifsym}

\begin{document}

\title{Human Activity Recognition with\\Convolutional Neural
Networks\thanks{This study was approved by the UCD Office of Research Ethics,
with authorization reference LS-17-107.}}

\author{\Letter Antonio Bevilacqua\inst{1} \and Kyle MacDonald\inst{2} \and Aamina
Rangarej\inst{2} \and Venessa Widjaya\inst{2} \and Brian Caulfield\inst{1} \and
Tahar Kechadi\inst{1}}

\authorrunning{A. Bevilacqua et al.}
\titlerunning{Human Activity Recognition with CNNs}

\tocauthor
\toctitle

\institute{Insight Centre for Data Analytics, UCD, Dublin,
  Ireland \and School of Public Health, Physiotherapy and Sports Science,
  UCD, Dublin, Ireland}

\maketitle

\begin{abstract}
  The problem of automatic identification of physical activities performed by
  human subjects is referred to as Human Activity Recognition (HAR). There exist
  several techniques to measure motion characteristics during these physical
  activities, such as Inertial Measurement Units (IMUs). IMUs have a cornerstone
  position in this context, and are characterized by usage flexibility, low
  cost, and reduced privacy impact. With the use of inertial sensors, it is
  possible to sample some measures such as acceleration and angular velocity of
  a body, and use them to learn models that are capable of correctly classifying
  activities to their corresponding classes. In this paper, we propose to use
  Convolutional Neural Networks (CNNs) to classify human activities. Our models
  use raw data obtained from a set of inertial sensors. We explore several
  combinations of activities and sensors, showing how motion signals can be
  adapted to be fed into CNNs by using different network architectures. We also
  compare the performance of different groups of sensors, investigating the
  classification potential of single, double and triple sensor systems. The
  experimental results obtained on a dataset of 16 lower-limb activities,
  collected from a group of participants with the use of five different sensors,
  are very promising.

\end{abstract}

\section{Introduction}
Human activity recognition (HAR) is a well-known research topic, that involves
the correct identification of different activities, sampled in a number of
ways. In particular, sensor-based HAR makes use of inertial sensors, such as
accelerometers and gyroscopes, to sample acceleration and angular velocity of a
body. Sensor-based techniques are generally considered superior when compared
with other methods, such as vision-based, which use cameras and microphones to
record the movements of a body: they are not intrusive for the users, as they do
not involve video recording in private and domestic context, less sensitive to
environmental noise, cheap and efficient in terms of power consumption
\cite{Cook2013,survey}. Moreover, the wide diffusion of embedded sensors in
smartphones makes these devices ubiquitous.

One of the main challenges in sensor-based HAR is the information
representation. Traditional classification methods are based on features that
are engineered and extracted from the kinetic signals. However, these features
are mainly picked on a heuristic base, in accordance with the task at
hand. Often, the feature extraction process requires a deep knowledge of the
application domain, or human experience, and still results in shallow features
only \cite{DBLP:journals/corr/abs-1305-0445}. Moreover, typical HAR methods
do not scale for complex motion patterns, and in most cases do not perform well
on dynamic data, that is, data picked from continuous streams.

On this regard, automatic and deep methods are gaining momentum in the field of
HAR. With the adoption of data-driven approaches for signal classification, the
process of selecting meaningful features from the data is deferred to the
learning model. In particular, CNNs have the ability to detect both spatial and
temporal dependencies among signals, and can effectively model scale-invariant
features \cite{7026300}.

In this paper, we apply convolutional neural networks for the HAR problem. The
dataset we collected is composed of 16 activities from the Otago exercise
program \cite{doi:10.1093/ageing/afq102}. We train several CNNs with signals
coming from different sensors, and we compare the results in order to detect the
most informative sensor placement for lower-limb activities. Our findings show
that, in most scenarios, the performance of a single sensor is comparable to the
performance of multiple sensors, but the usage of multiple sensor configurations
yields slightly better results. This suggests that collinearities exist among
the signals sampled with sensors on different placements.

The rest of the paper is organized as follows: Section \ref{section:related}
gives a brief overview of the state of the art of deep learning models for
activity recognition. Section \ref{section:methodology} presents our dataset,
the architecture of our neural network, and the methodology adopted in this
study. The experimental results are discussed in Section
\ref{section:results}. Some concluding remarks and future extensions for this
study are provided in Section \ref{section:conclusions}.

\section{Related works}
\label{section:related}
Extensive literature has been produced about sensor-based activity
recognition. Bulling \textit{et al.} \cite{Bulling:2014:THA:2578702.2499621}
give a broad introduction to the problem, highlighting the capabilities and
limitations of the classification models based on static and shallow
features. Alsheikh \textit{et al.}  \cite{DBLP:journals/corr/AlsheikhSNDLT15}
introduce a first approach to HAR based on deep learning models. They generate a
spectrogram image from an inertial signal, in order to feed real images to a
convolutional neural network. This approach overcomes the need for reshaping the
signals in a suitable format for a CNN, however, the spectrogram generation step
simply replaces the process of feature extraction, adding initial overhead to
the network training. Zeng \textit{et al.} \cite{7026300} use raw acceleration
signals as input for a convolutional network, applying 1-D convolution to each
signal component. This approach may result in loss of spatial dependencies among
different components of the same sensor. They focus on public datasets, obtained
mainly from embedded sensors (like smartphones), or worn sensors placed on the
arm. A similar technique is suggested by Yang \textit{et al.}
\cite{Yang:2015:DCN:2832747.2832806}. In their work, they use the same public
datasets, however, they apply 2-D convolution over a single-channel
representation of the kinetic signals. This particular application of CNNs for
the activity recognition problem is further elaborated by Ha \textit{et al.}
\cite{7727224}, with a multi-channel convolutional network that leverages both
acceleration and angular velocity signals to classify daily activities from a
public dataset of upper-limb movements. The classification task they perform is
personalized, so the signals gathered from each participant are used to train
individual learning models.

One of the missing elements in all the previously described contributions about
deep learning models is a comparison of the classification performance of
individual sensors or group of sensors. Our aim in this paper is to implement a
deep CNN that can properly address the task of activity recognition, and then
compare the results obtained with the adoption of different sensor
combinations. We also focus on a set of exercise activities that are part of the
Otago exercise program. To the best of our knowledge, this group of activities
has never been explored before in the context of activity recognition.

\section{Data and methodology}
\label{section:methodology}
The purpose of this paper is to assess the classification performance of
different groups of IMU sensors for different activities. We group the target
activities into four categories, and, for each category, we aim at identifying
the best placement for the inertial units, as well as the most efficient
combination of sensors, with respect to the activity classification task.

\subsection{Sensors and data acquisition}
Five sensors were used for the data collection phase. Each sensor is held in
place by a neoprene sleeve. For this study, we set the placement points as
follows:

\begin{itemize}
\item two sensors placed on the distal third of each shank (left and right),
  superior to the lateral malleolus;
\item two sensors centred on both left and right feet, in line with the head of
  the fifth metatarsal;
\item one sensor placed on the lumbar region, at the fourth lumbar vertebrae.
\end{itemize}

\begin{figure}
  \centering
  \includegraphics[width=\textwidth]{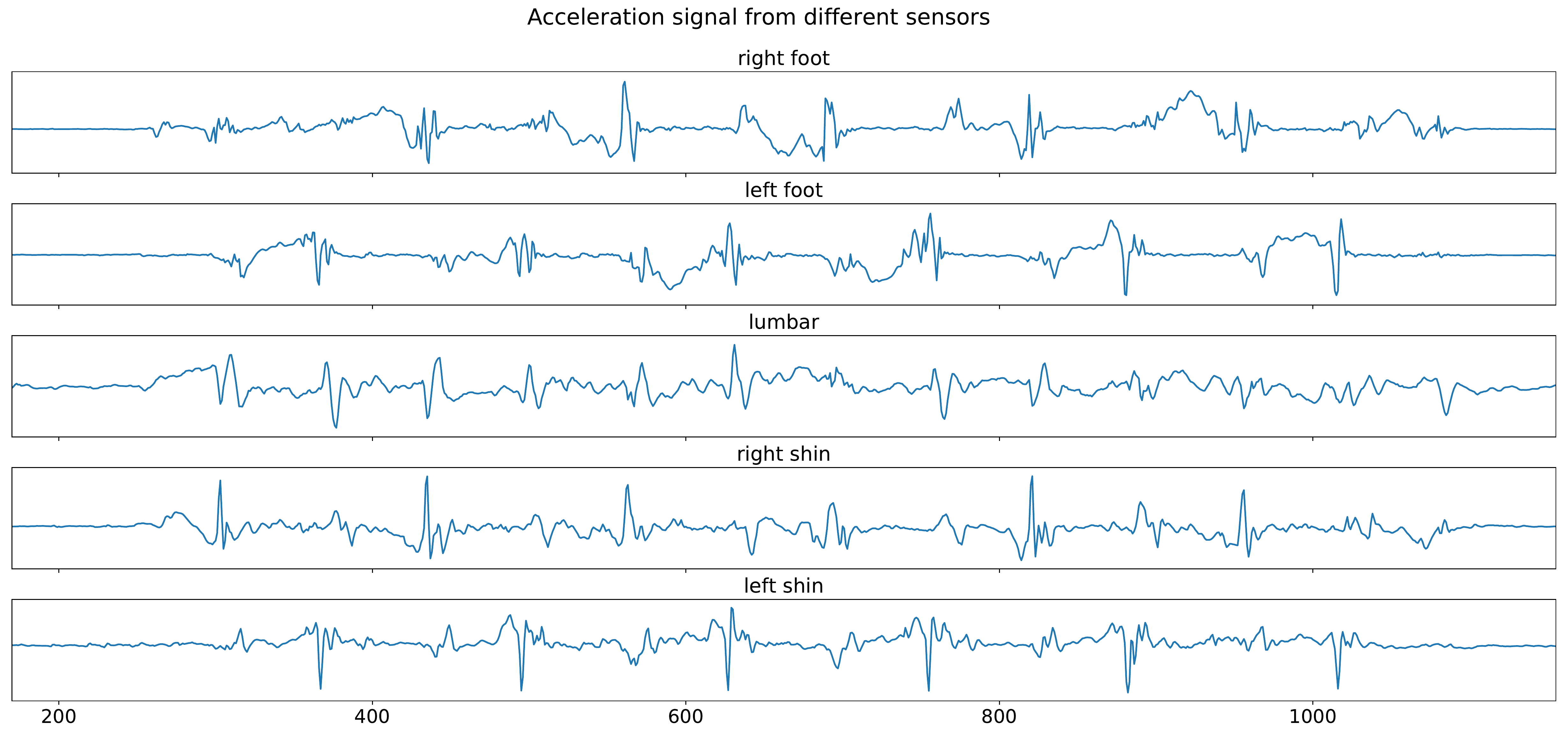}
  \caption{This acceleration $x$ components correspond to roughly 10 seconds of
    activity, acquired with the five sensors used in this study. Signals sampled
    by different sensors may show very discordant patterns and characteristics.}
  \label{pic:rawsignal}
\end{figure}

We explore three main sensor configurations: with a \textbf{single device}
setup, we classify the activity signal coming from each individual sensor. In
the \textbf{double device} setup, four combinations of two sensors are tested:
shin sensors (right and left), foot sensors (right and left), right sensors and
left sensors (foot and shin). When testing the \textbf{triple device} setups,
the lumbar sensor is included in each one of the double sensor configurations.

The chosen device for this study is Shimmer3 \cite{Burns2010}. The Shimmer3 IMU
contains a wide set of kinetic sensors, but we are interested in sampling
acceleration and angular velocity only. Both these quantities are captured by
triaxial sensors, so each Shimmer3 device returns a set of six signals (three
acceleration components, over the axes $x$, $y$ and $z$, and three angular
velocity components, over the same axes). The sampling rate is set to 102.4 Hz
for all sensors. The accelerometer is configured to have a range of $\pm2g$,
while the gyroscope range is set to 500 dps. We adopted this particular
configuration in order to avoid aliasing when sampling the target activities, as
gait-related human motion usually locates in the frequency range of 0.6-5.0 Hz
\cite{Godfrey}.

\subsection{Target activities and population}
The physical activities targeted in this paper are part of the Otago Exercise
Programme (OEP), a programme of activities designed to reduce the risk of
falling among the elderlies \cite{doi:10.1093/ageing/afq102}. In particular, we
grouped 16 different activities into four categories: \textbf{walk},
\textbf{walking balance}, \textbf{standing balance}, and \textbf{strength}. None
of the Otago warm-up activities is included in this study.

\begin{description}
\item[Walk]: it is composed of backwards walking (bawk), sideways walking
  (sdwk), walking and turning around (wktrn). These three activities have all
  wide and diverse range of movements, especially for the foot sensors.
\item[Walking Balance]: it is composed of heel to toe walking backwards
  (hetowkbk), heel walking (hewk), tandem walking (tdwk), toe walking
  (towk). These activities are based on similar ranges of movements.
\item[Standing Balance]: it is composed of single leg stance (sls), and tandem
  stance (tdst). The signals sampled from these two activities are mainly flat,
  as they require the subject to move only once in order to change the standing
  leg from left to right.
\item[Strength]: it is composed of knee extension (knex), knee flexion (knfx),
  hip abduction (hpabd), calf raise (cars), toe raise (tors), knee bend (knbn),
  and sit to stand (std). As some of these activities are performed by using
  each individual leg separately, all the sensor configurations involving both
  right and left sides are not applicable to this group.
\end{description}

A standard operating procedure defines the execution setup for all the target
activities, in terms of holding and pausing times, overall duration, starting
and ending body configurations. The same operating procedure is applied to all
the subjects involved in the study.

The group of 19 participants consists of 7 males and 12 females. Participants
have a mean age of 22.94$\pm$2.39, a mean height of 164.34$\pm$7.07 cm, and a
mean weight of 66.78$\pm$11.92 kg.

\subsection{Dataset}
\label{sub:dataset}
Once the signal is acquired from the activity, it is segmented into small
overlapping windows of 204 points, corresponding to roughly 2 seconds of
movements, with a stride of 5 points. A reduced size of the windows is generally
associated with a better classification performance \cite{banos_iwwbio_2014},
and in the context of CNNs, it facilitates the training process as the network
input has a contained shape. Therefore, each window comes in the form of a
matrix of values, of shape $6N \times 204$, where $N$ is the number of sensors
used to sample the window. The dense overlapping among windows guarantees high
numerosity of training and testing samples. As the activities have different
execution times, and different subjects may execute the same activity at
different paces, the resulting dataset is not balanced. The distributions of the
example windows over the activity classes for the five target groups are listed
in table \ref{table:labels}.

\begin{table}[]
\centering
\caption{Label distributions for the activity groups\\~}
\label{table:labels}
\begin{tabular}{|c|cc|c|c|}
\hline
\textbf{activity} & \textbf{windows} & \textbf{percentage} & \textbf{activity group}        & \textbf{total}         \\ \hline
bawk              & 19204            & 39.88               & \multirow{3}{*}{walk}          & \multirow{3}{*}{48143} \\
sdwk              & 22077            & 45.85               &                                &                        \\
wktrn             & 6925             & 14.38               &                                &                        \\ \hline
hetowkbk          & 4130             & 9.44                & \multirow{4}{*}{walk balance}  & \multirow{4}{*}{43754} \\
hewk              & 17796            & 40.67               &                                &                        \\
tdwk              & 4578             & 10.46               &                                &                        \\
towk              & 17250            & 39.42               &                                &                        \\ \hline
sls               & 20006            & 65.05               & \multirow{2}{*}{stand balance} & \multirow{2}{*}{30759} \\
tdst              & 10753            & 34.95               &                                &                        \\ \hline
knex              & 7500             & 12.14               & \multirow{7}{*}{strength}      & \multirow{7}{*}{76854} \\
knfx              & 6398             & 10.42               &                                &                        \\
hpabd             & 5954             & 9.62                &                                &                        \\
cars              & 6188             & 10.11               &                                &                        \\
tors              & 5815             & 9.41                &                                &                        \\
knbn              & 26452            & 34.42               &                                &                        \\
sts               & 8533             & 13.86               &                                &                        \\ \hline
\end{tabular}
\end{table}

For assessing the performance of our classification system, we use a classic
5-fold cross-validation approach. We partition the available datasets based on
the subjects rather than on the windows. This prevents overfitting over the
subjects and helps to achieve better generalisation results. In this regard, 4
participants out of 19 are always kept isolated for testing purposes, so each
fold is generated with an 80/20 split.

\subsection{Input adaptation and network architecture}
\begin{figure}
  \centering
  \includegraphics[width=\textwidth]{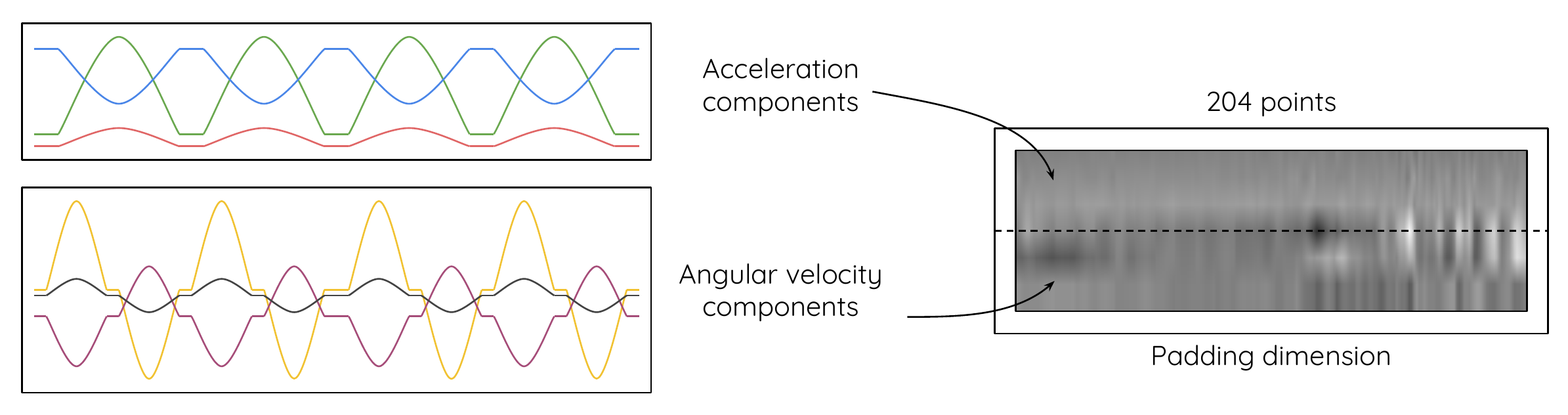}
  \caption{The signal components are stacked on top of each other to form a
    bidimensional matrix of values. Additional sensors would generate new
    channels of the same shape.} 
  \label{pic:input}
\end{figure}

The shape of the input examples that is fed to the network depends on the sensor
configuration, as each sensor samples 6 signal components that are then arranged
into a $6 \times 204$ single-channel, image-like matrix, as described in Section
\ref{sub:dataset}. Therefore, the input of the network has shape $6 \times 204
\times N$, where $N$ is the number of channels and is equal to the number of
sensors used for the sampling. This input adaptation is known as model-driven
\cite{survey}, and it is effective in detecting both spatial and temporal
features among the signal components \cite{7727224}. Figure \ref{pic:input}
shows how the signal components are stacked together and form the input image
for the network.

\begin{figure}
  \centering
  \includegraphics[width=\textwidth]{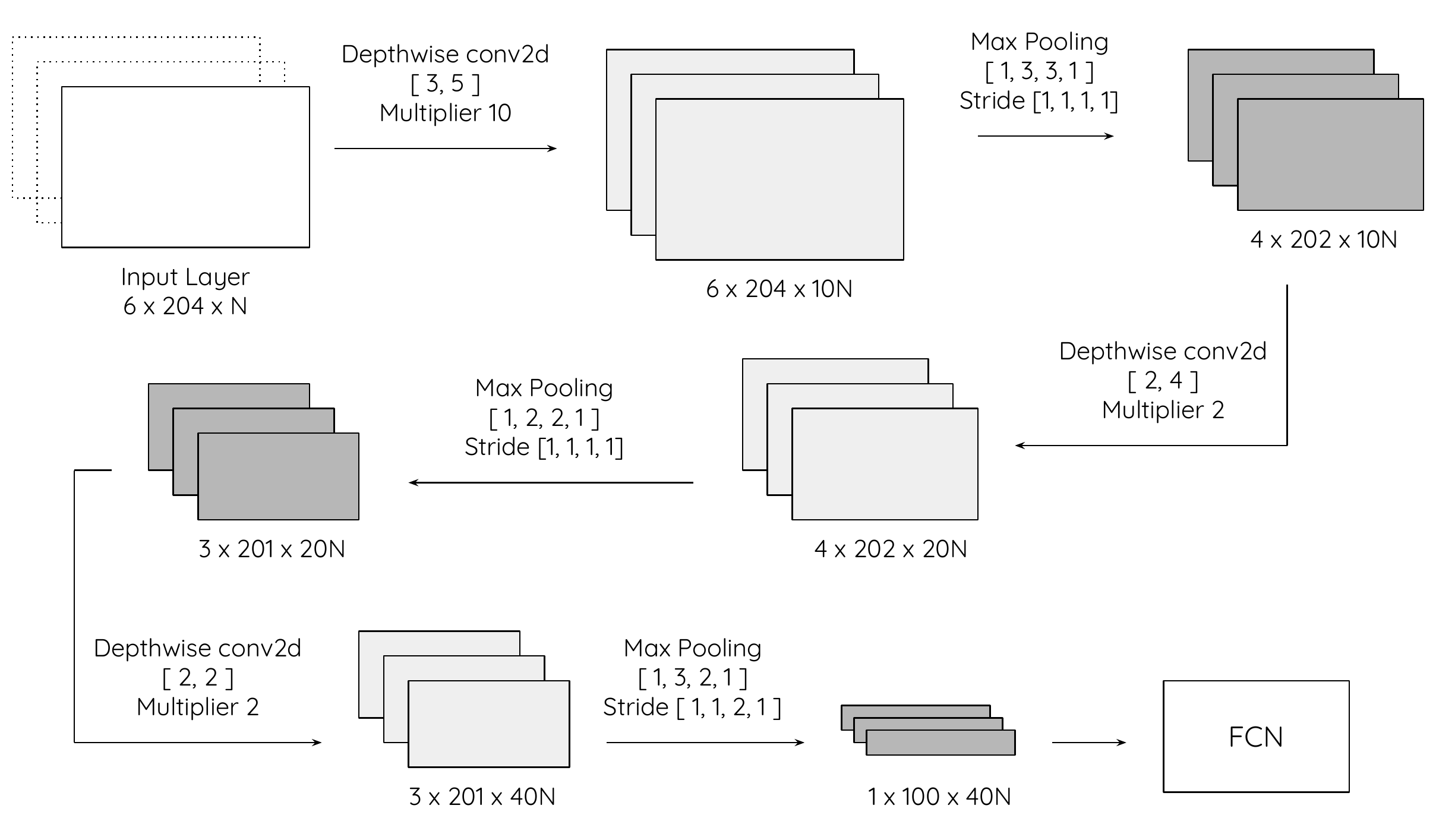}
  \caption{Our CNN architecture, where N represents the number of sensors used
    during the activity sampling. Regardless of the value of N, the network
    structure does not change, as depthwise convolution applies different
    filters to each one of the input channels.}
  \label{pic:architecture}
\end{figure}

The full structure of our convolutional model is shown in Figure
\ref{pic:architecture}. After the input layer, three convolutional layers
interleave with three max-pooling layers. The depthwise convolution operation
generates multiple feature maps for every input channel, with kernels of size $3
\times 5$, $2 \times 4$ and $2 \times 2$ in the first, second and third
convolutional layer respectively. The input of every convolutional layer is
properly padded so that no loss of resolution is determined from the convolution
operation. Batch normalization is applied after each convolutional layer. The
three max-pooling layers use kernels of size $3 \times 3$, $2 \times 2$ and $3
\times 2$ respectively. A fully connected network follows, composed of three
dense layers of 500, 250 and 125 units. The dense layers are regularized with
dropout during the training phase, with a 0.5 probability of keeping each
neuron. The ReLU function is used as activation function within the whole
network, while the loss is calculated with the cross entropy function. The Adam
optimizer is used as stochastic optimization method
\cite{DBLP:journals/corr/KingmaB14}. The output layer is composed ok $m$ units,
where $m$ corresponds to the number of activities in each group. The softmax
function will return the most likely class of the input windows in the
multi-class classification task.

We select a set of hyperparameters that are kept constant for all the activity
groups and sensor configurations, based on literature best practices
\cite{DBLP:journals/corr/abs-1206-5533} and empirical observations. We use a
batch size of 1024, as we find this value to speed up the learning process when
compared with smaller sizes, without being computationally too complex to
manage. The number of training epochs varies from 150 to up to 300, according to
the behaviour of individual configurations. The initial learning rate is fixed to
0.005. The network is implemented with the TensorFlow framework
\cite{tensorflow2015-whitepaper}, version 1.7. Our objective is not to build the
most performant network for the task, but it is rather to compare the
classification potential of different sensors. The rationale behind out
architectural choices relies therefore on a rather standard network
configuration, based on small kernels, standard regularization methods, and a
compact set of hyperparameters. In our experience, three convolutional layers
will lead to overfitting when no regularization method is applied. However,
introducing dropout stabilizes the learning, and the network performance does
not benefit from the inclusion of further convolutional or dense layers.

\section{Experimental results}
\label{section:results}
\begin{figure}
  \centering
  \includegraphics[width=\textwidth]{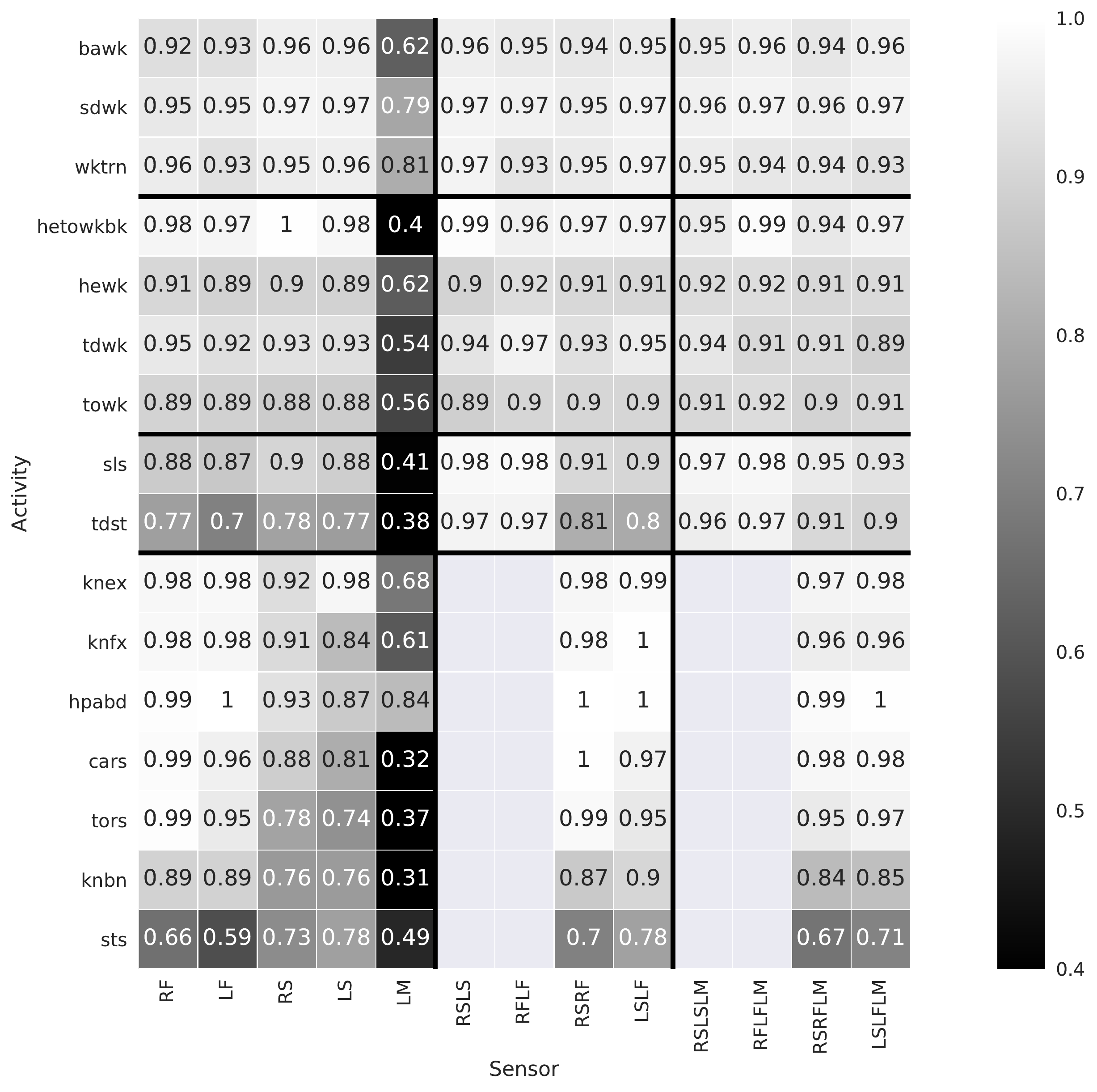}
  \caption{F-scores for the sensor configurations applied to the activity
    groups. Empty tiles for the strength group correspond to regions where
    particular sensor configurations were not applicable (asymmetric activities
    cannot be classified by using sensors on both the right and the left sides
    at the same time). The colour scheme spans from dark tones of black for low
    values of F-score to lighter tones for high F-score values. Lighter tiles
    denote better results than darker tiles. In order to emphasize small
    differences in the matrix, the minimum value for the tile colour gradient is
    set to be 0.4 (approximatively 0.1 higher than the smallest F-score value),
    so scores below this value will be marked with a black tile.}
  \label{pic:results}
\end{figure}

In order to evaluate our classifier, we collect individual precision and recall
scores for every combination of sensors and activities, and we then compute the
F-scores. A synoptic overview of the results is presented in Figure
\ref{pic:results}.

In the results shown in Figure \ref{pic:results}, the sensor combinations lay on
the $x$ axis. Starting from the left, there are right foot (RF), left foot (LF),
right shin (RS), left shin (LS), lumbar (LM), and all the other target setups
(for instance, the triple setup on the right side is indicated by RSRFLM, that
is, right shin, right foot, lumbar). The activities are arranged on the $y$
axis. Activity and sensor groups are indicated by the black grid of horizontal
and vertical lines. Each tile in the picture contains the F-score obtained for
the corresponding activity and sensor configuration. The colour gradient of the
tiles corresponds to the F-scores, and helps to identify high-level performance
for activity groups or sensor configurations.

The vertical strip of tiles corresponding to the lumbar sensor (LM) clearly
shows that this single sensor does not hold any significant discriminating
power, nor it adds meaningful information when included in the triple sensor
group, shown in the rightmost region of the picture. Overall, a strong pattern
on the sensor configurations does not appear to emerge: the units placed on the
feet show very similar results to the units placed on the shins, without clear
distinction in terms of discriminating power.

\begin{figure}
  \centering
  \includegraphics[width=.27\textwidth]{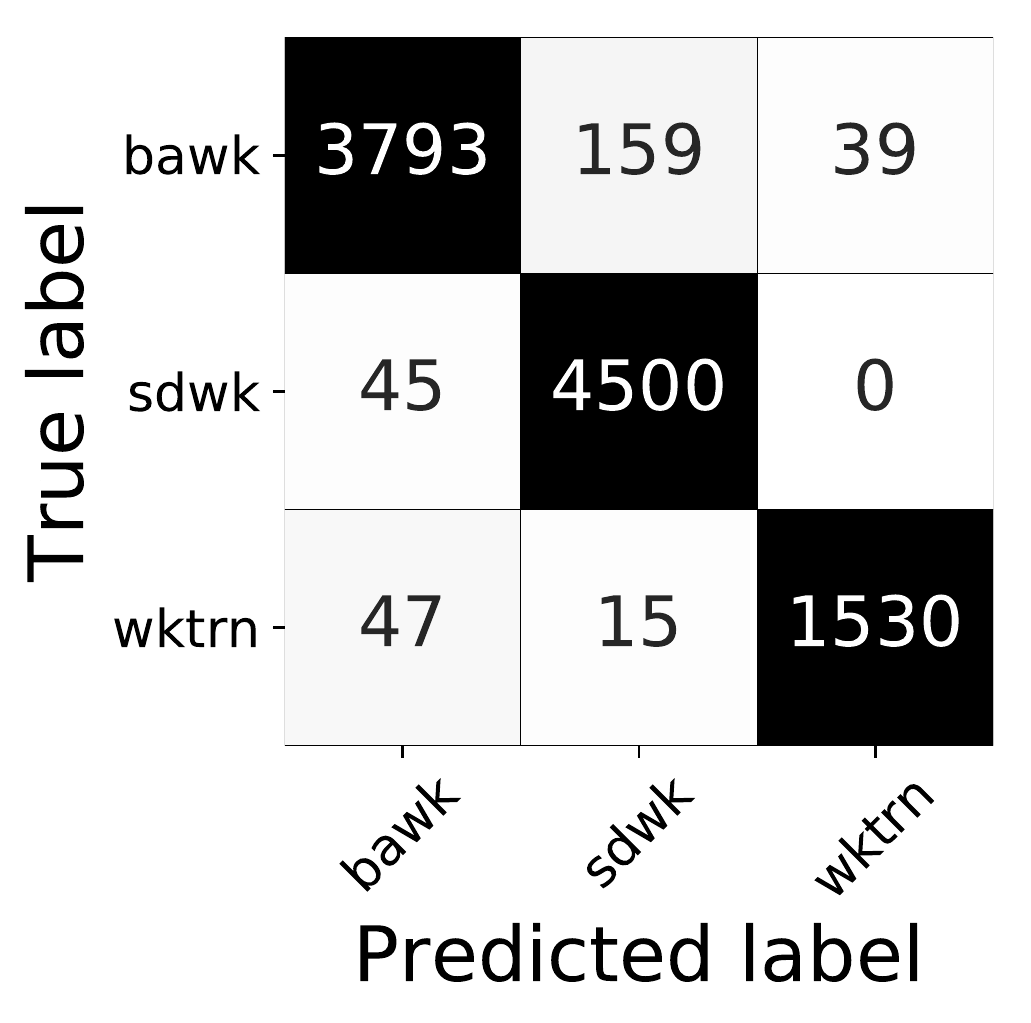}
  \caption{Confusion matrix for the walk group. The left shin sensor was used.}
  \label{pic:confwalk}
\end{figure}

\begin{figure}
  \centering
  \includegraphics[width=.33\textwidth]{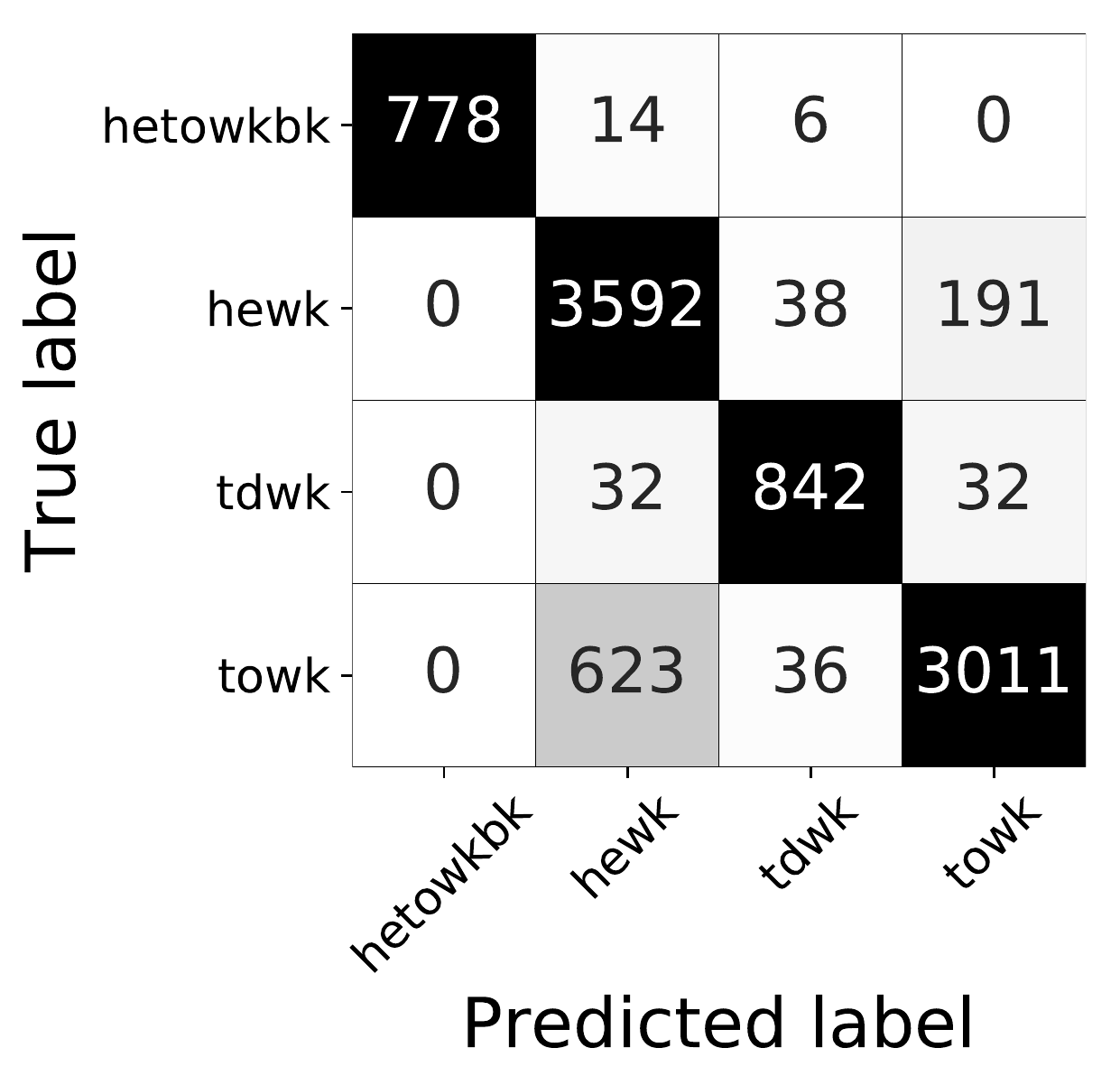}
  \caption{Confusion matrix for the walking balance group. The combination of
    right shin and right foot was used.}
  \label{pic:confwalkbalance}
\end{figure}

\begin{figure}
  \centering
  \includegraphics[width=.22\textwidth]{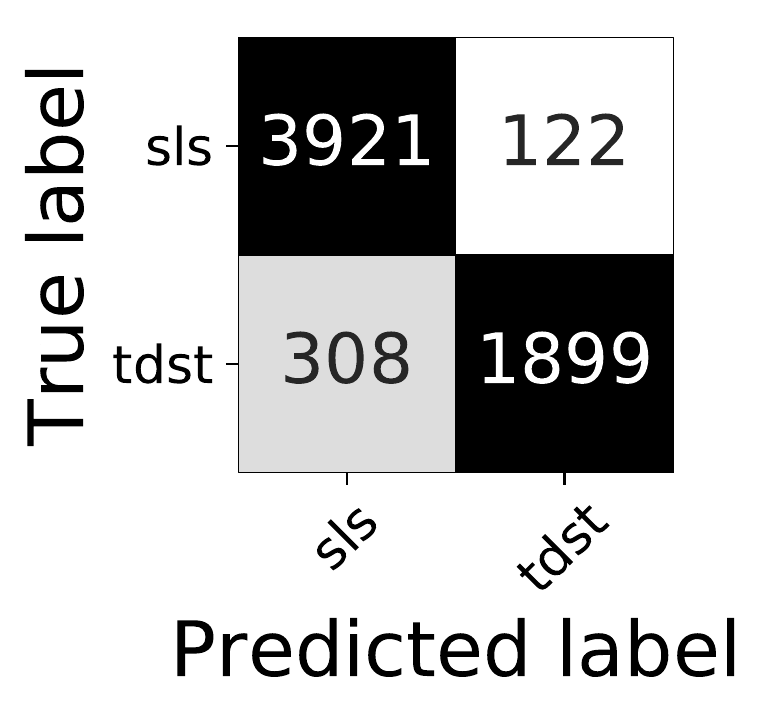}
  \caption{Confusion matrix for the standing balance group. The combination of
    right shin and right foot was used.}
  \label{pic:confstandbalance}
\end{figure}

\begin{figure}
  \centering
  \includegraphics[width=.5\textwidth]{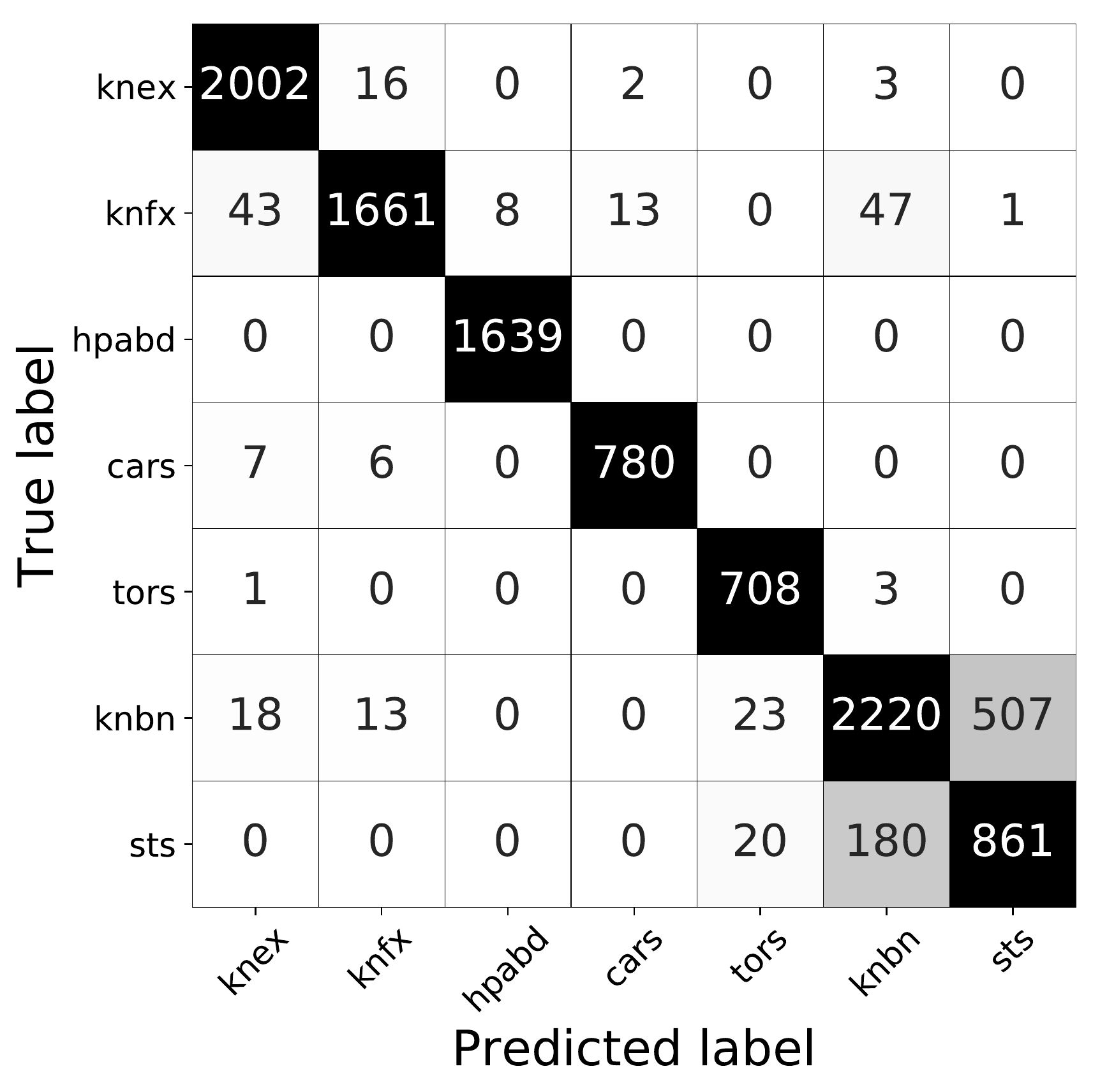}
  \caption{Confusion matrix for the strength group. The combination of left
    shin, left foot and lumbar sensors was used.}
  \label{pic:confstrength}
\end{figure}

The confusion matrices resulting from the evaluation process over the test
datasets are shown in Figures \ref{pic:confwalk}, \ref{pic:confwalkbalance},
\ref{pic:confstandbalance} and \ref{pic:confstrength}, for the walk group, the
walking balance group, the standing balance group, and the strength group
respectively. The colour gradient for each matrix is normalized on the rows.

The walk group scores interesting results for every sensor configuration. Single
sensors perform slightly worse than multiple sensor setups, however, there seems
to be no difference between two sensors and three sensors. From the confusion
matrix in Figure \ref{pic:confwalk}, we observe that the two majority classes,
bawk and sdwk, determined a reasonably limited amount of misclassified instances,
while the minority class, wktrn, only recorded 4\% of false negatives.

The same behaviour is shown for the walking balance group. In this case, the
hetowk and tdwk activities, which represent the 9.44\% and 10.46\% of the entire
group dataset respectively, performed remarkably well. For the first activity,
only 3\% of the instances were incorrectly classified, while the proportion of
misclassification for the second activity is 8\%. The confusion matrix in Figure
\ref{pic:confwalkbalance} indicates that the towk and hewk labels, the majority
classes, got a rate of false positives of 5\% and 17\% respectively, in favour
of one each other. From the global confusion matrix in Figure \ref{pic:results},
these two classes correspond to slightly darker bands within the activity
group. As defined in the exercise program, heel walking and toe walking present
some similarities.

The standing balance group, whose confusion matrix is reported in Figure
\ref{pic:confstandbalance}, was not properly classified with a single
sensor. The heavy class imbalance, in conjunction with the monotonicity of the
signals sampled from these two activities, skewed most of the misclassified
instances towards the majority class, sls, as indicated by the confusion
matrix in Figure \ref{pic:confstandbalance}. Nonetheless, the F-scores indicate
that symmetric combinations of sensors (right and left foot, right and left
shin) were able to discriminate between the two better than the asymmetric ones
(right side, left side).

As for the strength group, multiple sensor configurations increased the
classification score remarkably when compared with single sensor
configurations, in some cases reaching perfect classification for classes such
as hpabd, cars or knfx. The two classes that lowered the overall group
performance are knbn and sts, as shown in Figure \ref{pic:confstrength}. They
are based on very similar movements, so weak values of precision and recall are
somehow expected.

\section{Conclusions and Future works}
\label{section:conclusions}
In this paper, we presented a CNN model for the HAR problem. We focused on a set
of activities extracted from a common exercise program for fall prevention,
training our model data sampled from different sensors, in order to explore the
classification capabilities of each individual unit, as well as groups of
units. Our experimental results indicate that convolutional models can be used
to address the problem of activity recognition in the context of exercise
programs. In most cases, combinations of two or three sensors lead to better
results compared to the adoption of single inertial units.

Further work on the application of convolutional models to real-world data is
recommended. More activities could be included in the workflow, and different
aggregations on the activities can be tested. In particular, it is recommended
to diversify the population of participants, in order to validate the
classification mechanism to wider age groups. A proper campaign of
hyperparameter tuning should be carried over the same set of activities and
inertial units, in order to boost the classification performance and reduce the
complexity of both the training and inference phases. The very same network
structure could be redesigned in an optimized fashion for the task at hand, with
particular emphasis on the input adaptation step. As an example, shaping the
input in a single $6N \times 204$ could lead to interesting results, as more
complex kernels would allow the inclusion of features involving multiple
sensors.

\bibliographystyle{splncs04}
%% \bibliography{main}{}

\begin{thebibliography}{10}
\providecommand{\url}[1]{\texttt{#1}}
\providecommand{\urlprefix}{URL }
\providecommand{\doi}[1]{https://doi.org/#1}

\bibitem{tensorflow2015-whitepaper}
Abadi, M., Agarwal, A., Barham, P., Brevdo, E., Chen, Z., Citro, C., Corrado,
  G.S., Davis, A., Dean, J., Devin, M., Ghemawat, S., Goodfellow, I., Harp, A.,
  Irving, G., Isard, M., Jia, Y., Jozefowicz, R., Kaiser, L., Kudlur, M.,
  Levenberg, J., Man\'{e}, D., Monga, R., Moore, S., Murray, D., Olah, C.,
  Schuster, M., Shlens, J., Steiner, B., Sutskever, I., Talwar, K., Tucker, P.,
  Vanhoucke, V., Vasudevan, V., Vi\'{e}gas, F., Vinyals, O., Warden, P.,
  Wattenberg, M., Wicke, M., Yu, Y., Zheng, X.: {TensorFlow}: Large-scale
  machine learning on heterogeneous systems (2015),
  \url{https://www.tensorflow.org/}, software available from tensorflow.org

\bibitem{DBLP:journals/corr/AlsheikhSNDLT15}
Alsheikh, M.A., Selim, A., Niyato, D., Doyle, L., Lin, S., Tan, H.P.: Deep
  activity recognition models with triaxial accelerometers. CoRR
  \textbf{abs/1511.04664} (2015), \url{http://arxiv.org/abs/1511.04664}

\bibitem{banos_iwwbio_2014}
Banos, O., Galvez, J.M., Damas, M., Pomares, H., Rojas, I.: Evaluating the
  effects of signal segmentation on activity recognition. In: International
  Work-Conference on Bioinformatics and Biomedical Engineering, IWBBIO 2014.
  pp. 759--765 (2014)

\bibitem{DBLP:journals/corr/abs-1206-5533}
Bengio, Y.: Practical recommendations for gradient-based training of deep
  architectures. CoRR  \textbf{abs/1206.5533} (2012),
  \url{http://arxiv.org/abs/1206.5533}

\bibitem{DBLP:journals/corr/abs-1305-0445}
Bengio, Y.: Deep learning of representations: Looking forward. CoRR
  \textbf{abs/1305.0445} (2013), \url{http://arxiv.org/abs/1305.0445}

\bibitem{Bulling:2014:THA:2578702.2499621}
Bulling, A., Blanke, U., Schiele, B.: A tutorial on human activity recognition
  using body-worn inertial sensors. ACM Comput. Surv.  \textbf{46}(3),
  33:1--33:33 (Jan 2014). \doi{10.1145/2499621},
  \url{http://doi.acm.org/10.1145/2499621}

\bibitem{Burns2010}
Burns, A., Greene, B.R., McGrath, M.J., O'Shea, T.J., Kuris, B., Ayer, S.M.,
  Stroiescu, F., Cionca, V.: Shimmer\textsuperscript{TM} a wireless sensor
  platform for noninvasive biomedical research. IEEE Sensors Journal
  \textbf{10}(9),  1527 -- 1534 (2010). \doi{10.1109/JSEN.2010.2045498}

\bibitem{Cook2013}
Cook, D., Feuz, K.D., Krishnan, N.C.: Transfer learning for activity
  recognition: a survey. Knowledge and Information Systems  \textbf{36}(3),
  537--556 (Sep 2013). \doi{10.1007/s10115-013-0665-3},
  \url{https://doi.org/10.1007/s10115-013-0665-3}

\bibitem{Godfrey}
Godfrey, A., Conway, R., Meagher, D., ÓLaighin, G.: Direct measurement of
  human movement by accelerometry  \textbf{30},  1364--86 (01 2009)

\bibitem{7727224}
Ha, S., Choi, S.: Convolutional neural networks for human activity recognition
  using multiple accelerometer and gyroscope sensors. In: 2016 International
  Joint Conference on Neural Networks (IJCNN). pp. 381--388 (July 2016).
  \doi{10.1109/IJCNN.2016.7727224}

\bibitem{DBLP:journals/corr/KingmaB14}
Kingma, D.P., Ba, J.: Adam: {A} method for stochastic optimization. CoRR
  \textbf{abs/1412.6980} (2014), \url{http://arxiv.org/abs/1412.6980}

\bibitem{doi:10.1093/ageing/afq102}
Thomas, S., Mackintosh, S., Halbert, J.: Does the ‘otago exercise
  programme’ reduce mortality and falls in older adults?: a systematic review
  and meta-analysis. Age and Ageing  \textbf{39}(6),  681--687 (2010).
  \doi{10.1093/ageing/afq102}, \url{http://dx.doi.org/10.1093/ageing/afq102}

\bibitem{survey}
Wang, J., Chen, Y., Hao, S., Peng, X., Hu, L.: Deep learning for sensor-based
  activity recognition: {A} survey. CoRR  \textbf{abs/1707.03502} (2017),
  \url{http://arxiv.org/abs/1707.03502}

\bibitem{Yang:2015:DCN:2832747.2832806}
Yang, J.B., Nguyen, M.N., San, P.P., Li, X.L., Krishnaswamy, S.: Deep
  convolutional neural networks on multichannel time series for human activity
  recognition. In: Proceedings of the 24th International Conference on
  Artificial Intelligence. pp. 3995--4001. IJCAI'15, AAAI Press (2015),
  \url{http://dl.acm.org/citation.cfm?id=2832747.2832806}

\bibitem{7026300}
Zeng, M., Nguyen, L.T., Yu, B., Mengshoel, O.J., Zhu, J., Wu, P., Zhang, J.:
  Convolutional neural networks for human activity recognition using mobile
  sensors. In: 6th International Conference on Mobile Computing, Applications
  and Services. pp. 197--205 (Nov 2014).
  \doi{10.4108/icst.mobicase.2014.257786}

\end{thebibliography}

\end{document}